# TOGGLE: Temporal Logic-Guided Large Language Model Compression for Edge

Khurram Khalil, Khaza Anuarul Hoque
*Department of Electrical Engineering and Computer Science*
*University of Missouri-Columbia, USA*
{khurram.khalil, hoquek}@missouri.edu

*Abstract*—Large Language Models (LLMs) deliver exceptional performance across natural language tasks but demand substantial computational resources, limiting their deployment on resource-constrained edge devices. Existing compression techniques, such as quantization and pruning, often degrade critical linguistic properties and lack formal guarantees for preserving model behavior. We propose TOGGLE (Temporal Logic-Guided Large Language Model Compression), a novel framework that leverages *Signal Temporal Logic* (STL) to formally specify and enforce linguistic properties during compression. TOGGLE employs an STL robustness-guided Bayesian optimization to systematically explore layer-wise quantization and pruning configurations, generating compressed models that formally satisfy specified linguistic constraints without retraining or fine-tuning. Evaluating TOGGLE on four LLM architectures (GPT-2, DeepSeek-V2 7B, LLaMA 3 8B, and Mistral 7B), we achieve up to 3.3× reduction in computational costs (FLOPs) and up to a 68.8% reduction in model size while satisfying all linguistic properties. TOGGLE represents the first integration of formal methods into LLM compression, enabling efficient, verifiable deployment of LLMs on edge hardware.

## I. INTRODUCTION

Large Language Models (LLMs) have revolutionized natural language processing, demonstrating unprecedented capabilities across tasks such as text generation, reasoning, and complex problem-solving [1], [2]. However, deploying these models on resource-constrained edge devices poses significant challenges due to their escalating size and computational complexity. Recent LLMs have drastically increased in scale, from GPT-3's 175 billion parameters [1] to PaLM's 540 billion [3], and DeepSeek's 671 billion [4], with models like GPT-4 speculated to be even larger. The sheer magnitude of these models exceeds the computational and energy capacities of edge hardware. Although specialized accelerators and optimized inference engines have improved feasibility, the energy demands remain prohibitive for typical edge applications, which usually operate under severe parameter-space constraints [5]. Consequently, there is a growing need for compression techniques that preserve critical linguistic and reasoning capabilities while significantly reducing computational and energy footprints.

Current LLM compression strategies predominantly utilize quantization and pruning. Quantization methods reduce model precision, but uniform approaches like 4-bit or 8-bit quantization often degrade performance on tasks involving long-range dependencies or nuanced contextual coherence [6]. Mixed-precision quantization, which allocates variable bit-widths across layers, improves efficiency but introduces an immense combinatorial search space, growing as $B^{2L}$ for an $L$-layer model with $B$ bit-width options [7]. Similarly, pruning removes redundant attention heads or feed-forward neurons [8], yet aggressive pruning can undermine a model's sequential coherence and context handling [9]. Combining quantization with pruning exacerbates these challenges, exponentially enlarging the search space. Moreover, existing approaches typically rely on expensive retraining or knowledge distillation [10], which may be impractical in resource-limited scenarios [11]. Recent automated strategies based on reinforcement learning [12] or Bayesian optimization [13] still primarily depend on heuristic approaches without formal assurances for preserving essential linguistic properties like coherence, factual accuracy, or contextual consistency [14], [15], [16]. Additionally, these methods often overlook layer-specific sensitivities to compression, despite evidence suggesting different layers serve distinct linguistic purposes [17]. Thus, an automated, formally grounded compression approach that systematically addresses these issues is urgently needed.

To overcome these limitations, we introduce a novel framework, Temporal Logic-Guided Compression of Large Language Models (TOGGLE). TOGGLE utilizes Signal Temporal Logic (STL) [18]—a formal specification language—to define and preserve critical linguistic properties during compression. Leveraging STL, TOGGLE employs robustness-guided Bayesian optimization to systematically explore the joint quantization-pruning space (layer-wise bit-widths and pruning ratios), ensuring the resulting compressed models meet formally specified behavioral requirements. Additionally, TOGGLE supports runtime adaptability by dynamically controlling the trade-off between inference quality and energy efficiency through configurable operating modes. Our key contributions are as follows:

- We encode essential LLM properties—coherence, factual accuracy, long-range dependency, and contextual consistency—as STL specifications, enabling compressed models to meet fine-grained behavioral requirements.
- We develop a robustness-guided Bayesian optimization framework that leverages STL specifications to jointly optimize quantization and pruning, systematically exploring the compression space.
- We enable runtime control of inference quality, dynamically trading accuracy for energy efficiency across operating modes.
- We produce compressed LLMs without retraining or fine-tuning, minimizing deployment overhead, and validate TOGGLE's adaptability across diverse datasets for edge deployment.

*Key Results:* We rigorously evaluated TOGGLE using four diverse LLMs (GPT-2, DeepSeek-V2 7B, LLaMA 3 8B, and Mistral 7B) across relevant NLP evaluation datasets. By formalizing linguistic properties as STL specifications, our robustness-guided optimization framework successfully generated efficient compressed models without retraining. TOGGLE achieved substantial reductions in estimated computational cost, by up to approximately 3.3× compared to baseline models, while also realizing significant model compression, reducing model size by up to nearly 68.8%. To our knowledge, TOGGLE is the first framework that successfully integrated formal methods into LLM compression, enabling the systematic generation and deployment of efficient, formally verified LLMs on resource-constrained edge devices.

## II. PROBLEM FORMULATION

### A. Formalizing LLMs

Large Language Models (LLMs) are mainly built upon the Transformer architecture, which is known for revolutionizing Natural Language Processing (NLP) tasks such as text generation, summarization, and reasoning. The core innovation of the transformer lies in its attention mechanism, particularly multi-head attention, which prioritizes relevant input elements by computing attention scores across the sequence. For each input token, the model derives three parameters: Query ($Q$), Key ($K$), and Value ($V$), calculated through linear transformations applied to the input embeddings. Attention scores assess token importance, and masks can exclude specific tokens, such as padding. Each head processes distinct subsets of $Q$, $K$, and $V$, enabling parallel analysis of diverse token relationships. The outputs of individual heads are concatenated and linearly combined to produce the final attention output, enhancing the model's ability to capture complex dependencies.

To formalize the compression problem, we define the following variables and notations (summarized in Table I). Let $M_{\text{base}}$ represent the uncompressed base LLM and $L = \{l_1, l_2, \ldots, l_n\}$ be the set of layers in $M_{\text{base}}$, where $n$ is the total number of layers. The model parameters (weights and biases) are denoted collectively as $W$. Let $x = (x_1, x_2, \ldots, x_t)$ denote the sequence of input tokens processed by the LLM model up to generation step $t$, where each $x_i$ is a token from the vocabulary $\hat{V}$. We define $\mathcal{C}_{\text{components}}$ as the set of distinct parameter groups within each layer that can be targeted for compression. For instance, for architectures akin to GPT, $\mathcal{C}_{\text{components}}$ encompass attention projection mechanisms that integrate $Q, K,$ and $V$ transformations, attention output transformations, and feed-forward sub-layers. In contrast, for architectures similar to LLaMA, $\mathcal{C}_{\text{components}}$ include individual $Q, K,$ and $V$ transformations, attention output transformations, and extended feed-forward sub-layers with gating mechanisms, reflecting the architectural variations in how attention and feed-forward operations are structured. The main outputs generated during inference include next-token probability distributions over the vocabulary $\hat{V}$ at generation step $t$, denoted as $P_{\text{base}}(\cdot \mid x_{1:t-1})$ for the base model and $P_{\text{compressed}}(\cdot \mid x_{1:t-1})$ for the compressed model, attention maps for layer $l$ at generation step $t$, represented as $A^l_{\text{base}}(t)$ and $A^l_{\text{compressed}}(t)$, and hidden state embeddings at generation step $t$, denoted as $e_{\text{base}}(t)$ and $e_{\text{compressed}}(t)$. For each layer $l \in L$ and component $c \in \mathcal{C}_{\text{components}}$ within the layer, we define a compression configuration tuple $(b_{l,c}, p_{l,c})$, where $b_{l,c}$ specifies the bit-width for quantization and $p_{l,c}$ denotes the pruning ratio, which will be optimized to achieve efficient compression. The global compression configuration space $\hat{\mathcal{C}}$ is defined as $\hat{\mathcal{C}} = \prod_{l=1}^{o} \prod_{c \in \mathcal{C}_{\text{components}}} (\mathcal{B} \times \mathcal{P})$, where $\mathcal{B}$ represents a set of allowable bit-widths for quantization and $\mathcal{P}$ defines the range of permissible pruning ratios during compression. A specific configuration $\kappa \in \hat{\mathcal{C}}$ maps $M_{\text{base}}$ to a compressed model. Note, we use the notation $M_{\text{compressed}}$ to refer to the compressed LLM models in general and use the term $M_{\text{compressed}}(\kappa)$ to refer to a specific configuration $\kappa$ mapped to a compressed LLM model.

### B. Linguistic properties of LLMs

While LLMs exhibit a wide array of emergent linguistic properties, ranging from syntactic accuracy to stylistic nuance [19], [1], [20], this paper focuses on four key linguistic properties critical for reliable text generation: *sequential coherence*, *long-range dependencies*, *contextual consistency*, and *factual accuracy*. However, our proposed method is generic and thus can be easily applied for encoding and preserving other desirable properties during compression.

TABLE I: Summary of Key Variables and Notations

| Notation | Description |
|---|---|
| $M_{\text{base}}$ | Uncompressed base LLM model. |
| $M_{\text{compressed}}$ | Compressed base LLM model. |
| $\hat{V}$ | The vocabulary of tokens for the LLM. |
| $L = \{l_1, l_2, \ldots, l_n\}$ | The set of layers in the LLM model, with $n$ total layers. |
| $\mathcal{C}_{\text{components}}$ | The set of compressible components within each layer. |
| $W$ | The collection of all model parameters. |
| $\hat{\mathcal{C}}$ | The global compression configuration space, defined as $\prod_{l=1}^{n} \prod_{c \in \mathcal{C}_{\text{components}}} (\mathcal{B} \times \mathcal{P})$. |
| $\mathcal{B} = \{1, 2, 3, \ldots, B_{\max}\}$ | Allowable bit-widths for quantization; $B_{\max}$ is the maximum bit-width. |
| $\mathcal{P} = \{p_1, p_2, \ldots, p_m\}$ | Set of allowable discrete pruning ratios, where $p_i \in [0, P_{\max}]$, i.e. $\{0.0, 0.1, \ldots, P_{\max}\}$, and $P_{\max} \leq 1$. |
| $(b_{l,c}, p_{l,c})$ | Bit-width $b_{l,c} \in \mathcal{B}$ and pruning ratio $p_{l,c} \in \mathcal{P}$ applied to component $c$ in layer $l$. |
| $E(\kappa)$ | Estimated computational cost for configuration $\kappa$. |
| $T$ | Maximum context length supported by the LLM architecture. |
| $t$ | Discrete time step index during generation/evaluation. |
| $T'$ | Length of the specific evaluation trace/time horizon for STL property checking ($1 \leq T' \leq T$). |
| $S$ | Sequence length of the input tokens |

Sequential coherence ensures logical flow and grammatical correctness within the generated text, underpinning fluent communication [14], [1]. Long-range dependencies, enabled by attention mechanisms, allow LLMs to maintain semantic connections with distant parts of the input [21], [22]. Contextual consistency is crucial for coherent multi-turn dialogues, requiring sustained topic and persona [15], [23]. Factual accuracy ensures generated text aligns with known facts, avoiding hallucinations [16], [24]. These properties were chosen due to their established importance and known sensitivity to naive compression methods [14], [9], [15], [16]. Thus, preserving these behaviors is essential when compressing LLMs for practical deployment, requiring targeted rather than uniform compression approaches.

### C. Definitions

Before formulating the compression objectives, we establish key definitions related to LLM operations. These constructs are essential for formally specifying and evaluating linguistic properties during compression. Throughout these definitions, we use $M_{\text{model}}$ as a generic term referring to either $M_{\text{base}}$ or $M_{\text{compressed}}(\kappa)$.

**Definition 1** (Time Window). *During text generation, at each generation step $t \in \mathbb{N}$, a time window $TW(t, k)$ specifies a lookahead horizon of $k$ steps, defined as $TW(t, k) = \{t, t+1, \ldots, t+k\} \subset \mathbb{N}$. This window is bounded by the maximum context length $T$, i.e., $t + k \leq T$. The window $TW(t, k)$ delineates the interval over which formal properties are evaluated.*

**Definition 2** (Next-Token Probability Distribution). *Let $x_{1:t-1} = (x_1, x_2, \ldots, x_{t-1})$ be the sequence of tokens generated up to step $t - 1$. The next-token probability distribution at generation step $t$, denoted by $P_{model}(\cdot|x_{1:t-1})$, is a function mapping vocabulary tokens $v \in \hat{V}$ to their probabilities $P_{model}(x_t = v|x_{1:t-1})$. This distribution resides in the probability simplex over the vocabulary: $P_{model}(\cdot|x_{1:t-1}) \in \Delta^{|\hat{V}|-1}$. $P_{model}(\cdot|x_{1:t-1})$ can represent either base or compressed model.*

**Definition 3** (Attention Map). *For a given layer $l$ during the generation step $t$, the attention map $A^l_{model}(t)$ is a matrix $A^l_{model}(t) \in \mathbb{R}^{N \times N}$. Here, $N = \min(t - 1, T)$ represents the effective sequence length being processed at step $t$, where $t - 1$ is the number of preceding tokens and $T$ is the maximum context length supported by the model architecture. Each element $A^l_{model}(t)[p, q]$ (with $1 \leq p, q \leq N$) quantifies the attention score from query position $p$ to key position $q$. $A^l_{base}(t)$ and $A^l_{compressed}(t)$ denote the maps for the base and compressed models, respectively.*

**Definition 4** (Contextual Embedding). *The hidden state (or contextual embedding) at generation step t, denoted by $e_{model}(t) \in \mathbb{R}^{\hat{d}}$, is a vector representation capturing accumulated contextual information from the preceding token sequence $x_{1:t-1}$. Here, $\hat{d}$ is the hidden dimension of the model. $e_{base}(t)$ and $e_{compressed}(t)$ represent the embeddings for the base and compressed models.*

**Definition 5** (Computational Cost). *Let $M_{model}$ represent either the base model $M_{base}$ or a compressed model $M_{compressed}(\kappa)$. The computational cost of inference for $M_{model}$, denoted $E(M_{model})$, estimates the inference complexity, using Floating-Point Operations (FLOPs) required for processing a representative sequence length S. This cost serves as a proxy for energy efficiency. To formalize the cost calculation, let $\beta_{l,c}(M_{model})$ denote the effective bit-width and $\pi_{l,c}(M_{model})$ denote the effective pruning ratio utilized for component c in layer l during inference with model $M_{model}$. For the $M_{base}$, the effective bit-width $\beta_{l,c}(M_{base})$ corresponds to the model's original precision (e.g., 16 or 32 bits), and the effective pruning ratio $\pi_{l,c}(M_{base})$ is zero. For the $M_{compressed}(\kappa)$ derived from a specific configuration $\kappa$. Using these effective parameters, the estimated FLOPs are calculated as [25], [26]:*

$$E(M_{model}) = C \times \sum_{l \in L} \sum_{c \in \mathcal{C}_{components}} \left[ (1 - \pi_{l,c}(M_{model})) \times |W^{l,c}| \right. \\ \left. \times S \times \frac{\beta_{l,c}(M_{model})}{b_{ref}} \right] \quad (1)$$

*where $|W^{l,c}|$ is the number of parameters in the baseline component c of layer l, S is the input tokens sequence length, $b_{ref}$ is a reference bit-width (e.g., 16), and C is a constant factor (typically $C = 2$). The optimization objective is to minimize the cost of the compressed model $E(M_{compressed}(\kappa))$, which is denoted simply as $E(\kappa)$ (refer to Sec III-C for more information) in the problem formulation, where the dependence on $\kappa$ implies the use of its specific $b_{l,c}$ and $p_{l,c}$ values for $\beta_{l,c}$ and $\pi_{l,c}$ respectively in Eq. 1.*

**Definition 6** (Inference Signal). *Given an input prompt d and a model $M_{model}$, an inference signal $\sigma_{d,M_{model}}$ is a finite sequence of observations generated during $T'$ discrete steps of the inference execution, indexed over the time horizon $\{1, 2, \ldots, T'\}$. The signal maps each time step $t \in \{1, \ldots, T'\}$ to a vector $\sigma_{d,M_{model}}(t) \in \mathbb{R}^m$. The observations $\sigma_{d,M_{model}}(t)$ contain relevant state information corresponding to the $t^{th}$ step of the inference process (e.g., derived from the next-token distribution $P_{M_{model}}(\cdot|x_{1:t-1})$, the hidden state $e_{M_{model}}(t)$, or attention maps $A^l_{M_{model}}(t)$ calculated during step t). The dimension m depends on the specific variables monitored for the STL specifications, and the total duration $T'$ is the length of the specific evaluation trace ($T' \geq 1$).*

**Definition 7** (STL Robustness Degree). *Let $\hat{\varphi} = \{\varphi_1, \ldots, \varphi_n\}$ be a set of STL specifications representing desired linguistic properties. Given a specific specification $\varphi_i \in \hat{\varphi}$, an inference signal $\sigma_{d,M_{model}}$, where $d \in D$ and D is a representative evaluation dataset of input prompts, and a generation step t, the robustness degree $\rho(\varphi_i, \sigma_{d,M_{model}}, t)$ is a real-valued function that quantifies how well the inference signal $\sigma_{d,M_{model}}$ satisfies the specification $\varphi_i$ at step t [18], [27]. A positive robustness degree ($\rho > 0$) indicates satisfaction, a negative robustness degree ($\rho < 0$) indicates violation, and a robustness degree of zero indicates that the specification is marginally satisfied, meaning the signal lies exactly on the boundary of satisfaction. Let $\rho_{th} = (\rho_{th}(\varphi_1), \ldots, \rho_{th}(\varphi_n))$ be a vector of user-specified, non-negative thresholds, where each $\rho_{th}(\varphi_i)$ defines the minimum acceptable satisfaction margin for the corresponding linguistic property specification $\varphi_i$. The robustness degree is computed recursively based on the structure of the STL formula $\varphi_i$, using operations such as minimum for conjunctions and maximum for disjunctions [18], [27].*

These formal definitions provide the necessary components for encoding LLM linguistic properties using Signal Temporal Logic (STL) specification formalism and formulating the optimization problem.

*D. Problem Statement*

The problem statement of this paper can be formalized as follows.
**Given:** A base LLM model $M_{base}$, a discrete search space in the form of $\hat{\mathcal{C}}$, a computational cost function $E(\kappa)$ estimating the inference cost; a set of critical linguistic properties formalized as STL specifications $\hat{\varphi}$; a representative evaluation dataset D containing input prompts d; and a vector of non-negative robustness thresholds $\rho_{th}$ specifying the minimum acceptable satisfaction margin for each specification.

**Find:** An optimal compression configuration $\kappa^* \in \hat{\mathcal{C}}$ that produces a compressed model $M_{compressed}(\kappa^*)$, minimizing the computational cost $E(\kappa^*)$. The optimization problem is formally stated as:

$$\kappa^* = \arg\min_{\kappa \in \hat{\mathcal{C}}} E(\kappa) \quad (2)$$

$$\text{s.t.} \quad \rho(\varphi_i, \sigma_{d,M_{compressed}(\kappa)}, 0) \geq \rho_{th}(\varphi_i), \quad (3)$$
$$\forall \varphi_i \in \hat{\varphi}, \quad \forall d \in D$$

Solving this yields a compressed LLM optimized for efficiency under verifiable behavioral constraints defined by STL.

### III. METHODOLOGY

The proposed TOGGLE methodology in Figure 1 provides a structured framework for compressing LLMs while preserving essential linguistic properties. Given an uncompressed LLM model $M_{base}$, key linguistic properties formalized as *STL specifications* $\hat{\varphi}$, and associated non-negative robustness thresholds $\rho_{th}$, the process begins. The *Compression Configuration Space Exploration* phase systematically defines the search space $\hat{\mathcal{C}}$ of per-layer/per-component quantization bit-widths and pruning ratios. Then, the subsequent *Robustness-Guided Optimization* phase employs robustness analysis to evaluate candidate configurations $\kappa \in \hat{\mathcal{C}}$ against the STL constraints. This guides the search towards configurations that satisfy linguistic properties while minimizing computational cost $E(\kappa)$. Finally, the *Optimal Configuration Identification* process selects Pareto-optimal configurations, yielding compressed LLM models $M_{compressed}$ that adhere to the predefined linguistic constraints and offer substantial efficiency gains suitable for resource-constrained deployment. The details of these steps are described as follows.

*A. STL Specification of Linguistic Properties*

The TOGGLE framework leverages STL [28] to formalize critical linguistic properties in LLMs undergoing compression. We choose STL since it is widely known for its ability to reason about real-valued signals over time, provide quantitative robustness measures offering advantages over qualitative logics [18], and has been successfully deployed in various domains [29], [30]. Specifically, we utilize standard STL operators: *always* operator ($\Box_{[a,b]}\phi$, sometimes written as $\mathbf{G}_{[a,b]}\phi$), which asserts that property $\phi$ holds throughout the discrete time interval $[a,b]$, and the *conjunction* operator ($\phi_1 \wedge \phi_2$), requiring simultaneous satisfaction of multiple conditions. Properties are evaluated over inference signals $\sigma$ generated over a time horizon $[1, T']$. Further details on STL syntax and quantitative semantics can be found in [18], [27]. Below, we detail the STL formulations for four key linguistic properties identified in Section II-B.

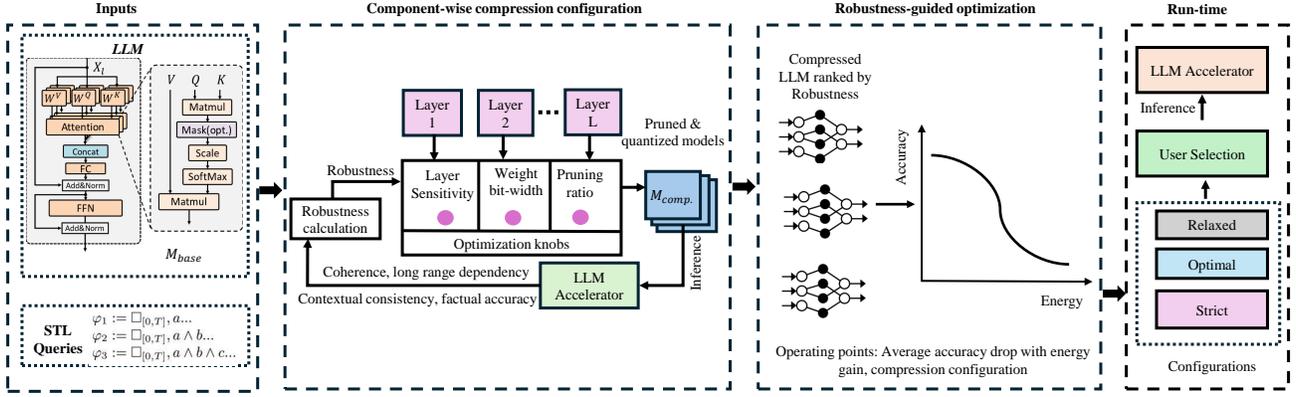

Fig. 1: Schematic of the proposed TOGGLE framework

Each STL specification $\varphi_i$ compares a relevant linguistic property against a user-defined predicate performance threshold (denoted as $\epsilon, \delta, \gamma, \tau$) to determine if the behavior is acceptable at each time step. The overall optimization constraint (Eq. 3) uses the quantitative robustness ($\rho$) — a measure of how strongly this threshold condition holds over the entire evaluation interval $[1, T']$ — and requires it to meet a minimum margin $\rho_{th}(\varphi_i)$. By setting this margin to zero ($\rho_{th} = 0$) in our work (Section IV-A (3)), the constraint $\rho \geq 0$ directly enforces that the conditions set by the internal Predicate Performance Thresholds must be satisfied throughout the evaluation period in the worst case.

*1) Sequential Coherence:* Sequential coherence requires that the next-token probability distributions generated by the compressed LLM remain close to those of the base model at each step, preserving local consistency. This can be measured using the Jensen-Shannon Divergence (JSD), a symmetric metric quantifying the similarity between probability distributions (see Definition 2), ensuring it stays below a specified threshold $\epsilon$, where $\epsilon \in (0, 1]$. Let $P_{b\_t} = P_{base}(\cdot|x_{1:t-1})$ and $P_{c\_t} = P_{compressed}(\cdot|x_{1:t-1})$. Then, sequential coherence can be defined as:

$$seq\_coh(P_{b\_t}, P_{c\_t}) := \epsilon - \mathrm{JSD}(P_{b\_t}, P_{c\_t}) \quad (4)$$

This can be formalized using STL specification, evaluated over the time horizon $T'$, as follows:

$$\varphi_1 := \Box_{[1,T']}(seq\_coh(P_{b\_t}, P_{c\_t}) \geq 0) \quad (5)$$

*2) Long-Range Dependencies:* Preserving long-range dependencies ensures that attention patterns, crucial for capturing distant relationships, are maintained after compression. We require the cosine similarity between the attention maps of each layer $l$ in the compressed and base models (Definition 3) to remain above a threshold $\delta$, where $\delta \in (0, 1]$. Let $\mathrm{sim}(A, B)$ denote cosine similarity between the flattened vector representations of matrices $A, B \in \mathbb{R}^{N \times N}$. Then, long-range dependency for layer $l$ can be defined as:

$$long\_dep_l(A^l_{base}(t), A^l_{compressed}(t)) := \mathrm{sim}(A^l_{base}(t), A^l_{compressed}(t)) - \delta \quad (6)$$

The comprehensive STL specification across all $\mathfrak{n}$ layers and the time horizon $T'$ is:

$$\varphi_2 := \bigwedge_{l=1}^{\mathfrak{n}} \Box_{[1,T']}(long\_dep_l(A^l_{base}(t), A^l_{compressed}(t)) \geq 0) \quad (7)$$

*3) Contextual Consistency:* Contextual consistency ensures that the semantic meaning captured by hidden state embeddings (Definition 4) is preserved during compression. We require the cosine similarity between embeddings from the compressed and base models to exceed a threshold $\gamma$ at each step, where $\gamma \in (0, 1]$. Then, contextual consistency can be defined as:

$$ctx\_cons(e_{base}(t), e_{comp}(t)) := \mathrm{sim}(e_{base}(t), e_{comp}(t)) - \gamma \quad (8)$$

This can be formalized using STL specification, evaluated over the time horizon $T'$, as follows:

$$\varphi_3 := \Box_{[1,T']}(ctx\_cons(e_{base}(t), e_{comp}(t)) \geq 0) \quad (9)$$

*4) Factual Accuracy:* Factual accuracy ensures the compressed model reliably assigns sufficient probability to known correct tokens (e.g., answers in QA, ground truth tokens). We require the ratio of the probability assigned to the correct token $y_{correct,t}$ by the compressed model versus the base model (using Definition 2) to stay above a threshold $\tau$, where $\tau \in (0, 1]$. Let $P_{base,t}(v) = P_{base}(v|x_{1:t-1})$ and $P_{comp,t}(v) = P_{compressed}(v|x_{1:t-1})$. Then, factual accuracy can be defined as:

$$fact\_acc(P_{base,t}, P_{comp,t}) := \frac{P_{comp,t}(y_{correct,t})}{P_{base,t}(y_{correct,t}) + \varepsilon_{div}} - \tau \quad (10)$$

where $\varepsilon_{div}$ is a small positive constant to prevent division by zero. This can be formalized using STL specification as:

$$\varphi_4 := \Box_{[1,T']}(fact\_acc(P_{base,t}, P_{comp,t}) \geq 0) \quad (11)$$

Through these STL specifications, TOGGLE provides a quantitative basis for evaluating linguistic property preservation during LLM compression, enabling the search for efficient models with formal behavioral guarantees. The chosen values for the predicate performance thresholds ($\epsilon, \delta, \gamma, \tau$) in each STL specification determine how closely the compressed model must replicate the baseline's behavior at each time step. For predicate thresholds, where higher values indicate better alignment, i.e., $\delta, \gamma, \tau$, thresholds closer to 1.0 impose stricter preservation requirements. In contrast, for $\epsilon$, lower values indicate higher similarity, and smaller thresholds enforce tighter constraints. These thresholds directly control the level of behavioral integrity enforced during the feasibility check.

### B. Compression Configuration

TOGGLE parameterizes compression across the layers ($L = \{l_1, \ldots, l_\mathfrak{n}\}$) and compressible components ($c \in \mathcal{C}_{components}$) of the base LLM $M_{base}$. This approach allows fine-grained control by assigning a specific quantization bit-width ($b_{l,c} \in \mathcal{B}$) and pruning ratio ($p_{l,c} \in \mathcal{P}$) to each component.

The bit-width $b_{l,c}$ controls the numerical precision of the parameters $W^{l,c}$ for component $c$ in layer $l$. TOGGLE employs specific quantization techniques tailored to the target precision level. For standard bit-widths (e.g., $b_{l,c} \geq 3$), the framework utilizes Learned Step-size Quantization (LSQ) [31]. For ultra-low precision ($b_{l,c} \leq 2$), the framework employs StretchedElasticQuant [32], a specialized method designed to better handle the dynamic range requirements of such low bit-widths.

The pruning ratio $p_{l,c}$ defines the fraction of the model parameters removed from component $c$ in layer $l$. TOGGLE implements unstructured magnitude-based pruning, setting weights with the lowest absolute values within each component $c$ in layer $l$ to zero independently to achieve the target sparsity level $p_{l,c}$ [33], [34].

A complete compression configuration $\kappa \in \hat{\mathcal{C}}$ constitutes a mapping that assigns a pair $(b_{l,c}, p_{l,c})$ to every targeted component $c$ in layer $l$ within the model:

$$\kappa = \{(l,c) \mapsto (b_{l,c}, p_{l,c}) \mid l \in L, c \in \mathcal{C}_{\text{components}}\} \quad (12)$$

This detailed configuration structure enables the sensitivity-aware, guided search performed by TOGGLE's optimization engine, necessitating an efficient strategy to navigate the vast configuration space.

### C. Computational Cost Estimation

To guide the optimization towards efficient models $M_{\text{compressed}}$, we estimate the computational cost of a compressed model $M_{\text{compressed}}(\kappa)$ using its configuration $\kappa$. Following standard practice [25], [26], we use Floating Point Operations (FLOPs) as the primary metric, which serves as a reliable method for computing energy consumption on target hardware.

Let $S$ be a representative input token sequence length used for estimation. The FLOPs for the baseline model ($M_{\text{base}}$, equivalent to $\kappa$ with full precision and zero pruning) are estimated as:

$$F_{\text{base}} \approx 2 \times \sum_{l \in L} \sum_{c \in \mathcal{C}_{\text{components}}} |W^{l,c}| \times S \quad (13)$$

where $|W^{l,c}|$ is the number of parameters in component $c$ in layer $l$. The factor of 2 accounts for multiply-accumulate operations.

For a compressed model specified by $\kappa = \{(b_{l,c}, p_{l,c})\}$, the estimated FLOPs ($F_{\text{compressed}}$) account for pruning and quantization:

$$F_{\text{compressed}}(\kappa) \approx 2 \times \sum_{l \in L} \sum_{c \in \mathcal{C}_{\text{components}}} \left[(1 - p_{l,c}) \times |W^{l,c}| \times S \times \frac{b_{l,c}}{b_{\text{ref}}}\right] \quad (14)$$

where $(1 - p_{l,c})$ reflects the reduction in active parameters due to pruning [35], and $(b_{l,c}/b_{\text{ref}})$ approximates the computational saving from using $b_{l,c}$-bit precision relative to a reference bit-width $b_{\text{ref}}$ (e.g., $b_{\text{ref}} = 16$) [25].

The computational cost function $E(\kappa)$ minimized in the optimization problem (Eq. 2) is directly proportional to this estimated FLOP count:

$$E(\kappa) \propto F_{\text{compressed}}(\kappa) \quad (15)$$

This FLOPs-based cost model provides a quantitative measure of computational efficiency used by the optimization algorithm.

### D. Robustness-Guided Optimization

The core of TOGGLE's optimization engine is designed to efficiently solve the constrained optimization problem defined in Section II-D. The goal is to navigate the high-dimensional, discrete compression configuration space $\hat{\mathcal{C}}$ to find the configuration $\kappa^*$ that minimizes the computational cost $E(\kappa)$ while satisfying the STL-defined linguistic properties. TOGGLE employs Bayesian Optimization (BO) [36] due to the high evaluation cost associated with STL robustness checking, which necessitates model instantiation and inference. BO internally handles discrete search variables (bit-widths, pruning ratios) using established techniques suitable for categorical and integer parameters. Specifically, the robustness-guided optimization methodology involves the following main steps:

**Step 1 (Candidate Configuration Proposal):** TOGGLE iteratively employs Gaussian Process (GP) surrogate models for the computational cost $E(\kappa)$ and minimum robustness scores $\rho_{min,i}(\kappa)$ across each linguistic property $\varphi_i$ for dataset $D$. Guided by the Expected Improvement under Constraints acquisition function [37], BO proposes the next configuration $\kappa_k$ to evaluate, balancing exploration with exploitation towards low-cost regions that satisfy all formal constraints defined by robustness thresholds $\rho_{th}(\varphi_i)$.

**Step 2 (Configuration Evaluation):** Each proposed candidate configuration $\kappa_k$ undergoes detailed evaluation. This includes estimating its computational cost $E(\kappa_k)$, instantiating the corresponding compressed model $M_{compressed}(\kappa_k)$, and performing inference over the evaluation dataset $D$. From this inference, signals $\sigma_{d, M_{compressed}(\kappa_k)}$ (as defined previously) are generated for each input prompt $d \in D$. Subsequently, the minimum robustness scores $\rho_{min,i}(\kappa_k)$ for each STL-defined linguistic property $\varphi_i$ are calculated across $D$, quantifying how strongly the compressed model satisfies each property specification.

**Step 3 (GP Model Update and Iteration):** The evaluated data tuple $(E(\kappa_k), \{\rho_{min,i}(\kappa_k)\}_{i=1}^n)$ from Step 2 is used to update the GP surrogate models. Steps 1 and 2 are iteratively repeated until the predefined evaluation budget is reached. Upon termination, all evaluated configurations that satisfy the formal feasibility condition, defined explicitly as:

$$\forall i \in \{1, \ldots, n\}: \quad \rho_{min,i}(\kappa) \geq \rho_{th}(\varphi_i), \quad (16)$$

are identified as *feasible* configurations. Analyzing these feasible points by plotting computational cost $E(\kappa)$ against overall minimum robustness $\rho_{min}(\kappa) = \min_i \rho_{min,i}(\kappa)$ reveals the achievable tradeoffs, forming an effective feasible *Pareto front*.

**Step 4 (Quantifying Average Property Preservation):** While the BO search is guided by STL robustness to ensure formal feasibility, practical operating modes are best defined using interpretable overall linguistic property preservation metrics. Thus, TOGGLE computes an Average Property Preservation (AvgPP) score for each feasible configuration $\kappa$, quantifying the level of preservation relative to the baseline model $M_{base}$. Specifically, AvgPP is computed as follows: First, for each linguistic property $\varphi_i$ and each input $d \in D$, the relevant underlying performance metric is computed for both $M_{base}$ and $M_{compressed}(\kappa)$ over the evaluation horizon $[1, T']$, yielding representative values $m_i(d, M_{base})$ and $m_i(d, M_{compressed}(\kappa))$, respectively. Next, a normalized preservation score $PS_i(d, \kappa)$ is calculated for each property and data point; this score reflects how well the compressed model retains the baseline's performance on that metric, using a normalization approach suitable for the metric's nature:

$$PS_i(d, \kappa) = \begin{cases} \max\left(0, 1 - \frac{m_i(d, M_{compressed}(\kappa))}{m_i(d, M_{base}) + \varepsilon_{norm}}\right), & \text{for } \varphi_1 \\ \min\left(1, \frac{m_i(d, M_{compressed}(\kappa))}{m_i(d, M_{base}) + \varepsilon_{norm}}\right), & \text{for } \varphi_2, \varphi_3, \varphi_4 \end{cases}$$

where $\varepsilon_{norm}$ is a small positive constant preventing division by zero. These scores are averaged across the dataset $D$ to obtain an average preservation score per property $i$, denoted as $\overline{PS_i}(\kappa) =$

$\text{avg}_{d \in D}[PS_i(d, \kappa)]$. Finally, AvgPP is computed by averaging these per-property scores:

$$AvgPP(\kappa) = \left(\frac{1}{n} \sum_{i=1}^{n} \overline{PS}_i(\kappa)\right) \times 100\%. \quad (17)$$

This AvgPP metric provides a single interpretable percentage score representing overall property preservation across specified linguistic dimensions.

**Step 5 (Operating Mode Selection):** Distinct operating modes are defined based on AvgPP targets. For illustrative purposes, we arbitrarily define three AvgPP-based operating modes: approximately 99% (**Strict**), 95% (**Optimal**), and 85% (**Relaxed**). However, TOGGLE can seamlessly accommodate any other user-specified AvgPP targets without fundamental methodological adjustments. To identify a representative configuration for each mode, TOGGLE first collects all feasible configurations satisfying $AvgPP(\kappa) \geq F_{target}$, where $F_{target}$ is the AvgPP level defining each mode. From this subset, the configuration with the lowest computational cost $E(\kappa)$ is selected. Thus, each mode represents the most computationally efficient solution found that satisfies both the formal feasibility constraint (Eq. 16) and the specified AvgPP target.

## IV. RESULTS AND ANALYSIS

This section presents and analyzes the empirical validation of the TOGGLE framework. We demonstrate its capability to compress diverse LLMs while ensuring adherence to formally specified linguistic properties via STL. We report and interpret the quantitative outcomes detailing compression efficacy, computational cost reduction (FLOPs), property preservation, and robustness for the selected operating modes derived from the Robustness-Guided Bayesian Optimization.

### A. Experimental Setup

*1) Models:* We evaluate TOGGLE on four pre-trained LLMs with diverse architectures and parameter scales: GPT-2 (124M parameters) [20], DeepSeek-V2 7B (7B parameters, a mixture-of-experts model) [38], LLaMA 3 8B (8B parameters, a dense transformer model optimized for research) [39], and Mistral 7B (7B parameters, a dense transformer model designed for efficiency in natural language tasks) [40]. This diverse model selection enables a comprehensive assessment of TOGGLE's effectiveness across varied architectural complexities and parameter scales.

*2) Datasets and STL Specifications:* To rigorously assess the preservation of STL-formalized linguistic properties while compressing LLMs, we employed a curated selection of standard NLP datasets, each tailored to evaluate a specific property. Specifically, for evaluating Sequential Coherence ($\varphi_1$), Long-Range Dependencies ($\varphi_2$), Contextual Consistency ($\varphi_3$), and Factual Accuracy ($\varphi_4$), we utilize LAMBADA [41], WikiText-2 [42], and TruthfulQA [43] datasets, respectively.

*3) STL Predicate Thresholds and Feasibility Definition:* Here, we define the thresholds used to establish formal feasibility within the Robustness-Guided Bayesian Optimization. Feasibility of a configuration $\kappa$ is determined by satisfying the constraint $\rho_{min,i}(\kappa) \geq \rho_{th}(\varphi_i)$ for all properties $i$. This involves internal predicate thresholds ($\epsilon, \delta, \gamma, \tau$) defining minimum acceptable instantaneous performance, and the robustness threshold $\rho_{th}(\varphi_i)$. For this study, the internal predicate thresholds were set aiming to preserve approximately 70% of the respective baseline linguistic behaviors, inspired by related work [44]. Specifically, we used $\delta = 0.70$ (for $\varphi_2$), $\gamma = 0.70$ (for $\varphi_3$), $\tau = 0.70$ (for $\varphi_4$), and $\epsilon = 0.25$ (for $\varphi_1$). With these defined, the minimum acceptable robustness threshold was set to zero for all properties: $\rho_{th}(\varphi_i) = 0$ for $i = 1, \ldots, n$. Consequently, the feasibility condition $\rho_{min,i}(\kappa) \geq 0$ directly enforces that the internal predicate thresholds (e.g., sim $\geq 0.70$, $JSD \leq 0.25$) are met in the worst case. The BO search seeks low-cost configurations within this formally defined feasible space.

TABLE II: Compression efficiency across operating modes for all evaluated models. BMS and CMS represent baseline and compressed model sizes (in megabytes). BF/T and CF/T refer to baseline and compressed FLOPs per token, measured in gigaflops (GFLOPs).

| Model | Mode | BMS | CMS | BF/T | CF/T |
|---|---|---|---|---|---|
| GPT-2 | Strict | | 210.8 | | 0.08 |
| | Optimal | 248 | 139.4 | 0.10 | 0.05 |
| | Relaxed | | 96.9 | | 0.04 |
| DeepSeek-V2 7B | Strict | | 11368 | | 9.5 |
| | Optimal | 14000 | 7266 | 12.4 | 5.9 |
| | Relaxed | | 4900 | | 4.1 |
| LLaMA 3 8B | Strict | | 14400 | | 13.3 |
| | Optimal | 16000 | 9600 | 14.6 | 8.1 |
| | Relaxed | | 6496 | | 5.6 |
| Mistral 7B | Strict | | 10934 | | 9.5 |
| | Optimal | 14000 | 6566 | 12.4 | 5.4 |
| | Relaxed | | 4368 | | 3.8 |

*4) Optimization Execution and Mode Selection:* The Robustness-Guided Bayesian Optimization was executed individually for each evaluated model to identify low-cost configurations satisfying the formal feasibility constraint. The optimization search space included quantization bit-widths $\mathcal{B} = \{2, 3, \ldots, 16\}$, which were selected based on typical quantization levels used in literature and practical hardware considerations for efficient deployment, and pruning ratios $\mathcal{P} = \{0.0, 0.1, \ldots, 0.5\}$, chosen to explore a broad yet practically feasible range of sparsity levels without compromising model functionality. The Bayesian Optimization was executed with a budget of 200 iterations per model to explore the high-dimensional search space adequately. After completing optimization, the resulting set of feasible configurations was collected. These configurations served as input for selecting representative operating modes based on predefined AvgPP targets as described in Section III-D.

*5) Implementation Details:* Experiments were conducted using PyTorch 2.1.0, CUDA 12.1, the RTAMT library [27] for STL robustness monitoring, and a BO implementation based on BoTorch [45]. The optimization and evaluation for all models were performed on a cluster node equipped with 4 NVIDIA A100 80GB GPUs. The total computation time for the BO search across all four LLMs amounted to **approximately 360 GPU hours**.

### B. Analysis of Efficiency Gains and Compression Strategies

This section analyzes the efficiency gains achieved by TOGGLE in terms of computation and model size reduction across the different operating modes. The core results are presented in Table II, which reports the baseline and compressed model sizes (BMS, CMS) and the estimated baseline and compressed giga FLOPs per token (BF/T, CF/T), respectively. Table II shows significant reductions in estimated FLOPs per token across all models and modes. For instance, the base LLaMA 3 8B model requires approximately 14.6 GFLOPs/token, which TOGGLE reduces to 8.1 GFLOPs/token in the Optimal mode, representing a **1.8× reduction in computation**, and further down to 5.6 GFLOPs/token in the Relaxed mode, implying a **2.6× FLOPs reduction**. Furthermore, the results in Table II also show substantial model size reductions achieved by the different operating modes. For instance, the base DeepSeek-V2 model (approx. 14000 MB) was compressed to **4900 MB** in the Relaxed mode, representing a significant **65% decrease in model size**. Similarly, the balanced Optimal mode reduced its size considerably to approximately 7200 MB, implying **48.1% compression ratio**. The compression ratio represents the percentage reduction in compressed model size achieved compared to the original baseline model. Similar significant reductions in absolute size were observed across all evaluated models, scaling with the aggressiveness of the selected operating mode (Strict resulting in the largest compressed size, Relaxed the smallest). The specific compression strategies underlying these size reductions are detailed in Table III, reporting the average quantization bit-width (AvgBits), average pruning ratio (AvgPrun), and resulting compression ratio (CR%) and FLOPS reduction (×) for each mode. For instance, the Relaxed mode for Mistral 7B combined an average bit-width of 7.0 and 40.0% average pruning to achieve the **highest observed compression ratio of 68.8%** that resulted in 3.3× reduction in FLOPs. As expected, the data confirms that higher AvgPP (i.e., stricter preservation of linguistic properties) necessitated more conservative compression strategies (higher AvgBits, lower AvgPrun), whereas the Relaxed mode utilized the most aggressive settings found within the feasible space. This demonstrates the BO's effectiveness in tailoring the fine-grained compression parameters $(b_{l,c}, p_{l,c})$ to achieve the target property preservation level while satisfying the underlying formal robustness constraints.

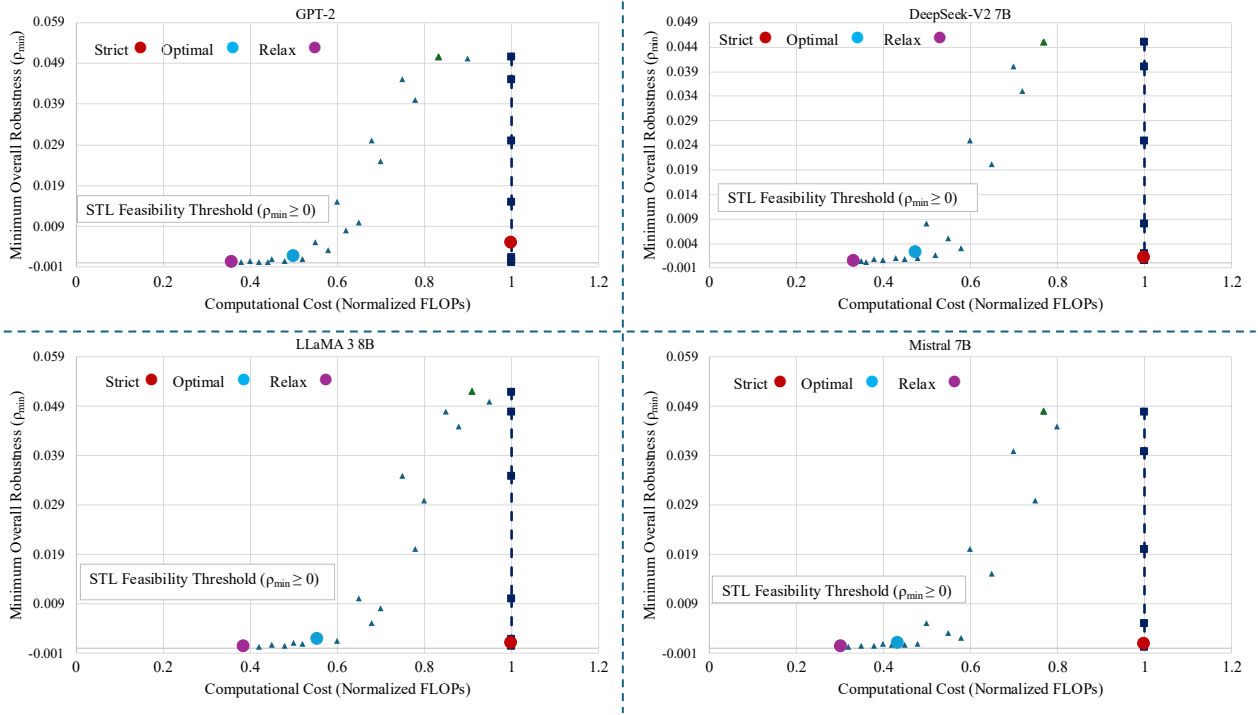

Fig. 2: Feasible Pareto fronts illustrating the trade-off between computational cost and minimum overall STL robustness ($\rho_{min}$) score.

### C. Pareto Front Analysis and Trade-off Interpretation

This section analyzes the Pareto front discovered through BO and explicitly characterizes the trade-off between computational efficiency and formal robustness. Figure 2 illustrates the candidate configurations identified by TOGGLE for all models, plotted in the space of computational cost versus minimum overall STL robustness. The feasible points found during the search are shown primarily as triangles, with additional feasible points near the baseline shown as squares. For instance, let us take the Mistral 7B model as an illustrative example—though similar trends are consistently observed for other models—the Pareto front reveals a steep trade-off near the Strict mode, meaning that minor robustness improvements incur disproportionately large increases in computational cost. Conversely, substantial efficiency gains can be achieved near the Optimal mode region with only minimal relaxation in robustness. The highlighted feasible Pareto fronts, implied by the boundary formed by the triangles and squares, in each subplot represent the optimal boundary between cost and formal robustness discovered by TOGGLE. The overlays of the selected Strict, Optimal, and Relaxed modes confirm their positions as efficient points relative to this underlying formal landscape, validating that the AvgPP-based selection aligns well with the cost-robustness trade-offs identified by the optimization. Specifically, this Pareto-front analysis reveals the cost of achieving higher formal robustness guarantees: significant computational cost increases may yield relatively small robustness gains at higher robustness levels (near the Strict mode), whereas substantial efficiency gains can be achieved with only minor relaxations in robustness near the Optimal mode. This analysis emphasizes TOGGLE's ability to systematically identify efficient and formally verified configurations, enabling informed decision-making between computational efficiency and formal behavioral guarantees.

Figure 3 presents a detailed analysis of property preservation percentages across different modes and models. Taking GPT-2 as an example—though similar trends are observed consistently across all evaluated models—the preservation profiles corresponding to each compression mode (Strict, Optimal, and Relaxed) demonstrate distinct characteristics. Specifically, the Strict mode achieves uniformly high preservation (98–99%) across all linguistic properties, clearly reflecting its conservative compression strategy. In contrast, the Optimal mode achieves balanced preservation close to the 95% target, while the Relaxed mode exhibits more varied property degradation. These distinct preservation profiles of the compression modes underscore the importance of employing multi-property STL specifications, as different linguistic properties exhibit varying degrees of sensitivity to compression, particularly at lower preservation targets.

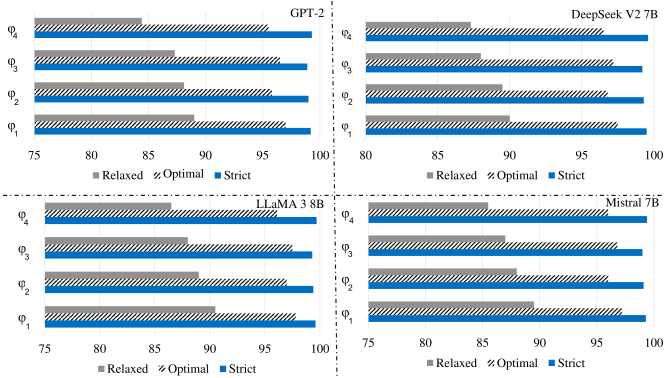

Fig. 3: Linguistic property preservation scores across different operating modes and models, normalized to 100% at baseline.

*D. Sensitivity Analysis of Predicate Performance Thresholds*

This section investigates the impact of varying predicate performance thresholds ($\epsilon, \delta, \gamma, \tau$) and reports their impact on FLOPs reduction and compression ratio. Due to the space constraints, we perform sensitivity analysis using the Mistral 7B model. However, we believe that this trend can be generalized to other LLMs as well. For the thresholds $\delta, \gamma$ and $\tau$, we vary them from 0.5 to 0.9, while for $\epsilon$ we vary from 0.15 to 0.35. We vary these thresholds one at a time while keeping the other thresholds fixed. For each setting, we identify the most computationally efficient configuration satisfying the corresponding feasibility constraint. Table IV reports the resulting FLOPs Reduction Factor (FR) and Compression Ratio (CR), showing how each threshold influences the achievable energy efficiency.

We observe that relaxing $\delta$ from its baseline of 0.7 to 0.5 yields the most significant efficiency gain, increasing the FR from 2.3× to approximately 3.1× and the CR from 53.1% to over 66%. Conversely, tightening this requirement to $\delta = 0.9$ significantly reduces efficiency, e.g., FR decreases to 1.5×, and CR decreases to 35.2%. Relaxing the thresholds $\gamma$ and $\epsilon$ also results in notable efficiency gains, though less pronounced than for $\delta$. Varying the $\tau$ exhibits the least impact on overall efficiency, suggesting it is potentially less constraining than other thresholds for this model. This analysis confirms that these thresholds provide effective control over the LLM compression outcome. It highlights that threshold related to attention mechanisms (i.e., $\delta$) pose significant bottlenecks for efficiency, and relaxing these specifically can yield substantial efficiency gains, albeit potentially at the cost of that particular property's preservation. Understanding these sensitivities is crucial for setting appropriate thresholds based on application-specific priorities for different linguistic behaviors.

TABLE III: Compression configuration parameters and resulting efficiency metrics for Optimal (O), Strict (S), and Relaxed (R) operating modes. AvgBits denotes the average quantization bit-width; AvgPrun is the average pruning ratio (%); CR is the overall compression ratio (%); and FR indicates the FLOPs reduction factor (×).

| Model | Mode | AvgBits | AvgPrun (%) | CR (%) | FR (×) |
|---|---|---|---|---|---|
| GPT-2 | S | 14.2 | 5.0 | 15.0 | 1.2 |
| | O | 10.0 | 20.0 | 43.8 | 2.0 |
| | R | 8.0 | 30.0 | 60.9 | 2.8 |
| DeepSeek-V2 7B | S | 14.0 | 10.0 | 18.8 | 1.3 |
| | O | 9.0 | 25.0 | 48.1 | 2.1 |
| | R | 7.0 | 35.0 | 65.0 | 3.0 |
| LLaMA 3 8B | S | 15.0 | 5.0 | 10.0 | 1.1 |
| | O | 10.0 | 10.0 | 40.0 | 1.8 |
| | R | 8.0 | 25.0 | 59.4 | 2.6 |
| Mistral 7B | S | 13.0 | 15.0 | 21.9 | 1.3 |
| | O | 8.0 | 20.0 | 53.1 | 2.3 |
| | R | 7.0 | 40.0 | 68.8 | 3.3 |

TABLE IV: FLOPs Reduction (FR) and Compression Ratio (CR) for different values of predicate performance thresholds. For each row, the boldfaced value indicates the threshold being varied, while the others are held constant.

| $\epsilon$ | $\delta$ | $\gamma$ | $\tau$ | FR (×) | CR (%) |
|---|---|---|---|---|---|
| **0.15** | 0.7 | 0.7 | 0.7 | 2.0 | 48.5 |
| **0.20** | 0.7 | 0.7 | 0.7 | 2.1 | 50.8 |
| **0.25** | 0.7 | 0.7 | 0.7 | 2.3 | 53.1 |
| **0.30** | 0.7 | 0.7 | 0.7 | 2.4 | 56.0 |
| **0.35** | 0.7 | 0.7 | 0.7 | 2.5 | 58.2 |
| 0.25 | **0.5** | 0.7 | 0.7 | 3.1 | 66.5 |
| 0.25 | **0.6** | 0.7 | 0.7 | 2.7 | 60.1 |
| 0.25 | **0.7** | 0.7 | 0.7 | 2.3 | 53.1 |
| 0.25 | **0.8** | 0.7 | 0.7 | 1.9 | 46.0 |
| 0.25 | **0.9** | 0.7 | 0.7 | 1.5 | 35.2 |
| 0.25 | 0.7 | **0.5** | 0.7 | 2.9 | 63.0 |
| 0.25 | 0.7 | **0.6** | 0.7 | 2.5 | 57.5 |
| 0.25 | 0.7 | **0.7** | 0.7 | 2.3 | 53.1 |
| 0.25 | 0.7 | **0.8** | 0.7 | 2.1 | 49.8 |
| 0.25 | 0.7 | **0.9** | 0.7 | 1.7 | 42.1 |
| 0.25 | 0.7 | 0.7 | **0.5** | 2.6 | 59.0 |
| 0.25 | 0.7 | 0.7 | **0.6** | 2.4 | 55.8 |
| 0.25 | 0.7 | 0.7 | **0.7** | 2.3 | 53.1 |
| 0.25 | 0.7 | 0.7 | **0.8** | 2.2 | 51.0 |
| 0.25 | 0.7 | 0.7 | **0.9** | 2.1 | 49.5 |

## V. CONCLUSION

This paper presented *TOGGLE*, a framework that leverages Signal Temporal Logic (STL) to compress Large Language Models (LLMs) while formally preserving critical linguistic properties—*sequential coherence*, *long-range dependencies*, *contextual consistency*, and *factual accuracy*—without requiring retraining. TOGGLE employs STL-guided Bayesian Optimization to find per-layer quantization and pruning configurations that satisfy formal behavioral constraints. We evaluated TOGGLE on four representative LLMs: GPT-2, DeepSeek-V2 7B, LLaMA 3 8B, and Mistral 7B. Results demonstrate that TOGGLE achieves up to 3.3× FLOPs reduction and up to 68.8% model size reduction, while maintaining positive STL robustness across all properties, thereby ensuring formal feasibility. In the future we plan to extend TOGGLE to multi-modal foundation models and incorporate hardware-aware optimization objectives, such as memory footprint and inference latency.

## VI. ACKNOWLEDGMENTS

This material is based upon work supported by the National Science Foundation (NSF) under Award Numbers: CCF-2323819. Any opinions, findings, conclusions, or recommendations expressed in this publication are those of the authors and do not necessarily reflect the views of the NSF.

## REFERENCES

[1] T. Brown, B. Mann, N. Ryder, M. Subbiah et al., "Language models are few-shot learners," *Advances in Neural Information Processing Systems*, vol. 33, pp. 1877–1901, 2020.

[2] A. Chowdhery, S. Narang, J. Devlin, M. Bosma, G. Mishra, A. Roberts, P. Barham, H. W. Chung, C. Sutton et al., "Palm: Scaling language modeling with pathways," *arXiv preprint arXiv:2204.02311*, 2022.

[3] S. Smith, M. Patwary, B. Norick, P. LeGresley, S. Rajbhandari, J. Casper, Z. Liu, S. Prabhumoye, G. Zerveas, V. Korthikanti et al., "Using deepspeed and megatron to train megatron-turing nlg 530b, a large-scale generative language model," *arXiv preprint arXiv:2201.11990*, 2022.

[4] D.-A. et al., "Deepseek llm: Scaling open-source language models with longtermism," 2024. [Online]. Available: https://arxiv.org/abs/2401.02954

[5] M. Zhou, K. Xu, and Z. Lin, "Efficient edge deployment of compressed llms," *Transactions on Machine Learning Research*, vol. 6, no. 3, pp. 1–25, 2023.

[6] T. Dettmers, M. Lewis, Y. Belkada, and L. Levy, "Llm.int8(): 8-bit matrix multiplication for transformers at scale," *arXiv preprint arXiv:2208.07339*, 2022.

[7] G. Xiao, W. Zhang, and Y. Liu, "Smoothquant: Accurate and efficient post-training quantization for large language models," *arXiv preprint arXiv:2301.12345*, 2023.

[8] E. Frantar and D. Alistarh, "Sparsegpt: Efficient pruning for large language models," *arXiv preprint arXiv:2302.12345*, 2023.

[9] X. Ma, Z. Li, and Y. Chen, "Sparse transformer architectures: Balancing efficiency and performance," *Proceedings of the 2023 International Conference on Learning Representations*, 2023.

[10] G. Hinton, O. Vinyals, and J. Dean, "Distilling the knowledge in a neural network," *arXiv preprint arXiv:1503.02531*, 2015.

[11] A. Gupta, R. Kumar, and V. Singh, "Compressing large language models for edge deployment: Challenges and opportunities," *Proceedings of the 2023 International Conference on Edge Computing*, pp. 456–467, 2023.

[12] Z. Yao, Y. Chen, and W. Zhang, "Compression of large language models using reinforcement learning," *arXiv preprint arXiv:2303.12345*, 2023.

[13] Y. Chen, L. Zhang, and X. Wang, "Bayesian optimization for neural network compression," *Proceedings of the 2023 AAAI Conference on Artificial Intelligence*, pp. 1234–1245, 2023.

[14] X. Liu, J. Zhang, and Y. Wang, "Coherence in large language models: Challenges and solutions," *Journal of Machine Learning Research*, vol. 24, no. 1, pp. 1–45, 2023.

[15] P. Henderson, W. Li, and Y. Chen, "Contextual consistency in multi-turn dialogues: A study on large language models," *Proceedings of the 2022 Conference on Empirical Methods in Natural Language Processing*, pp. 1234–1245, 2022.

[16] Z. Lin, M. Zhao, and K. Xu, "Factual accuracy in large language models: A systematic evaluation," *Transactions on Machine Learning Research*, vol. 5, no. 2, pp. 1–30, 2023.

[17] H. Lin, X. Wang, and Z. Li, "Layer-wise sensitivity analysis for transformer-based models," *Advances in Neural Information Processing Systems*, vol. 35, 2022.

[18] A. Donzé and O. Maler, "Robust satisfaction of temporal logic over real-valued signals," in *Proceedings of the 8th International Conference on Formal Modeling and Analysis of Timed Systems*, ser. FORMATS'10. Berlin, Heidelberg: Springer-Verlag, 2010, p. 92–106.

[19] R. Bommasani, D. A. Hudson, E. Adeli *et al.*, "On the opportunities and risks of foundation models," *arXiv preprint arXiv:2108.07258*, 2021.

[20] A. Radford, J. Wu, R. Child, D. Luan, D. Amodei, and I. Sutskever, "Language models are unsupervised multitask learners," *OpenAI Blog*, 2019, available at: https://openai.com/research/language-models-are-unsupervised-multitask-learners.

[21] A. Vaswani, N. Shazeer, N. Parmar, J. Uszkoreit, L. Jones, A. N. Gomez, Ł. Kaiser, and I. Polosukhin, "Attention is all you need," *Advances in neural information processing systems*, vol. 30, 2017.

[22] J. Devlin, M.-W. Chang, K. Lee, and K. Toutanova, "BERT: Pre-training of deep bidirectional transformers for language understanding," *Proceedings of the 2019 Conference of the North American Chapter of the Association for Computational Linguistics: Human Language Technologies*, vol. 1, pp. 4171–4186, 2019.

[23] X. L. Li and P. Liang, "Prefix-tuning: Optimizing continuous prompts for generation," *Proceedings of the 59th Annual Meeting of the Association for Computational Linguistics*, pp. 4582–4597, 2021.

[24] T. Zhang, Y. Zhang, Q. Liu, Z. Ma, and W. Y. Wang, "Faithful or extractive? on mitigating the faithfulness-abstractiveness trade-off in abstractive summarization," *Proceedings of the 2022 Conference on Empirical Methods in Natural Language Processing*, pp. 3456–3468, 2022.

[25] Y. Cheng, D. Wang, P. Zhou, and T. Zhang, "Efficient deep learning: A survey on making deep learning models smaller, faster, and better," *ACM Computing Surveys*, vol. 55, no. 12, pp. 1–34, 2023.

[26] J. Hoffmann, S. Borgeaud, A. Mensch, E. Buchatskaya, T. Cai, E. Rutherford, D. de Las Casas, L. A. Hendricks, J. Welbl, A. Clark *et al.*, "Training compute-optimal large language models," *arXiv preprint arXiv:2203.15556*, 2022.

[27] D. Nickovic and T. Yamaguchi, "Rtamt: Online robustness monitors from stl," 2020. [Online]. Available: https://arxiv.org/abs/2005.11827

[28] O. Maler and D. Nickovic, "Monitoring temporal properties of continuous signals," *Formal Techniques, Modelling and Analysis of Timed and Fault-Tolerant Systems*, pp. 152–166, 2004.

[29] O. A. Beg, L. V. Nguyen, T. T. Johnson, and A. Davoudi, "Signal temporal logic-based attack detection in dc microgrids," *IEEE Transactions on Smart Grid*, vol. 10, no. 4, pp. 3585–3595, 2019.

[30] E. Bartocci, J. Deshmukh, A. Donzé, G. Fainekos, O. Maler, D. Ničković, and S. Sankaranarayanan, "Specification-based monitoring of cyber-physical systems: a survey on theory, tools and applications," *Lectures on Runtime Verification: Introductory and Advanced Topics*, pp. 135–175, 2018.

[31] S. K. Esser, J. L. McKinstry, D. Bablani, R. Appuswamy, and D. S. Modha, "Learned step size quantization," 2020. [Online]. Available: https://arxiv.org/abs/1902.08153

[32] Z. Liu, C. Zhao, H. Huang, S. Chen, J. Zhang, J. Zhao, S. Roy, L. Jin, Y. Xiong, Y. Shi, L. Xiao, Y. Tian, B. Soran, R. Krishnamoorthi, T. Blankevoort, and V. Chandra, "Paretoq: Scaling laws in extremely low-bit llm quantization," 2025. [Online]. Available: https://arxiv.org/abs/2502.02631

[33] S. Han, H. Mao, and W. J. Dally, "Deep compression: Compressing deep neural networks with pruning, trained quantization and huffman coding," in *International Conference on Learning Representations (ICLR)*, 2016, arXiv preprint arXiv:1510.00149 (2015). [Online]. Available: https://arxiv.org/abs/1510.00149

[34] T. Gale, E. Elsen, and S. Hooker, "The state of sparsity in deep neural networks," *arXiv preprint arXiv:1902.09574*, 2019. [Online]. Available: https://arxiv.org/abs/1902.09574

[35] T. Hoefler, D. Alistarh, T. Ben-Nun, N. Dryden, and A. Peste, "Sparsity in deep learning: Pruning and growth for efficient inference and training in neural networks," 2021. [Online]. Available: https://arxiv.org/abs/2102.00554

[36] J. Snoek, H. Larochelle, and R. P. Adams, "Practical bayesian optimization of machine learning algorithms," in *Advances in Neural Information Processing Systems*, vol. 25, 2012, pp. 2951–2959.

[37] J. R. Gardner, M. J. Kusner, Z. E. Xu, K. Q. Weinberger, and J. P. Cunningham, "Bayesian optimization with inequality constraints," in *Proceedings of the 31st International Conference on Machine Learning (ICML)*, ser. Proceedings of Machine Learning Research, E. P. Xing and T. Jebara, Eds., vol. 32. PMLR, 22–24 Jun 2014, pp. 937–945. [Online]. Available: https://proceedings.mlr.press/v32/gardner14.html

[38] DeepSeek-AI, "Deepseek-v2: A strong, economical, and efficient mixture-of-experts language model," 2024, available at: https://arxiv.org/abs/2405.04434.

[39] A. G. et al., "The llama 3 herd of models," 2024. [Online]. Available: https://arxiv.org/abs/2407.21783

[40] A. Q. Jiang, A. Sablayrolles, A. Mensch, C. Bamford, D. S. Chaplot, D. de las Casas, F. Bressand, G. Lengyel, G. Lample, L. Saulnier, L. R. Lavaud, M.-A. Lachmann, P. Tallec, T. Carbonneau, T. Hassen, and T. Lavril, "Mistral 7b," 2023, available at: https://arxiv.org/abs/2310.06825.

[41] D. Paperno, G. Kruszewski, A. Lazaridou, N. Q. Pham, R. Bernardi, S. Pezzelle, M. Baroni, and R. Fernández, "The lambada dataset: Word prediction requiring a broad discourse context," vol. 1, pp. 1525–1534, 2016, available at: https://aclanthology.org/P16-1144.

[42] S. Merity, C. Xiong, J. Bradbury, and R. Socher, "Pointer sentinel mixture models," 2016, available at: https://arxiv.org/abs/1609.07843.

[43] S. Lin, J. Hilton, and O. Evans, "Truthfulqa: Measuring how models mimic human falsehoods," 2022, available at: https://arxiv.org/abs/2109.07958.

[44] O. Spantidi and I. Anagnostopoulos, "Automated energy-efficient dnn compression under fine-grain accuracy constraints," in *2023 Design, Automation Test in Europe Conference Exhibition (DATE)*, 2023, pp. 1–6.

[45] M. Balandat, B. Karrer, D. R. Jiang, S. Daulton, B. Letham, A. G. Wilson, and E. Bakshy, "BoTorch: A Framework for Efficient Monte-Carlo Bayesian Optimization," in *Advances in Neural Information Processing Systems 33*, 2020. [Online]. Available: http://arxiv.org/abs/1910.06403